\documentclass[3p,times]{elsarticle}

\makeatletter
\def\ps@pprintTitle{%
 \let\@oddhead\@empty
 \let\@evenhead\@empty
 \def\@oddfoot{\centerline{\thepage}}%
 \let\@evenfoot\@oddfoot}
\makeatother

\usepackage{lmodern}

\usepackage{amssymb}

\usepackage{lineno,hyperref}

\usepackage{times}
\usepackage{latexsym}

\usepackage{CJKutf8}
\usepackage[utf8]{inputenc}
\usepackage[T1]{fontenc}

\usepackage{microtype}

\usepackage{tabularx}
\usepackage{multirow}
\usepackage{graphicx}
\usepackage{subfigure}
\usepackage{amsmath}
\usepackage{amssymb}
\usepackage{booktabs}
\usepackage[linesnumbered,ruled,vlined]{algorithm2e} 

\SetAlFnt{\small}
\SetAlCapFnt{\small}
\SetAlCapNameFnt{\small}
\SetAlCapHSkip{0pt}
\IncMargin{-\parindent}
\usepackage{algorithmic}
\usepackage{color, colortbl}
\definecolor{LightCyan}{rgb}{0.88,1,1}
\definecolor{Gray}{gray}{0.9}

\usepackage{tikz}

\usepackage[misc]{ifsym}

\begin{document}

\begin{frontmatter}




\title{QBSUM: a Large-Scale Query-Based Document Summarization Dataset from Real-world Applications}


\author[UofA]{Mingjun Zhao\corref{EqualContrib}}
\author[Tencent]{Shengli Yan\corref{EqualContrib}}
\cortext[EqualContrib]{Equal contribution}
\author[UofA,UofM]{Bang Liu\corref{CorrespondingAuthor}}
\cortext[CorrespondingAuthor]{Corresponding Author}\ead{bang.liu@umontreal.ca}

\author[Tencent]{Xinwang Zhong}
\author[Tencent]{Qian Hao}
\author[Tencent]{Haolan Chen}
\author[UofA]{Di Niu}
\author[Tencent]{Bowei Long}
\author[Tencent]{Weidong Guo}

\address[UofA]{University of Alberta, 116 St. and 85 Ave., Edmonton, AB T6G 2R3, Canada}
\address[Tencent]{Platform and Content Group, Tencent, 10000 Shennan Ave, Shenzhen 518057, China}
\address[UofM]{University of Montreal, 2900 Edouard Montpetit Blvd, Montreal, Quebec H3T 1J4, Canada}


\begin{abstract}
Query-based document summarization aims to extract or generate a summary of a document which directly answers or is relevant to the search query.
It is an important technique that can be beneficial to a variety of applications such as search engines, document-level machine reading comprehension, and chatbots.
Currently, datasets designed for query-based summarization are short in numbers and existing datasets are also limited in both scale and quality.
Moreover, to the best of our knowledge, there is no publicly available dataset for Chinese query-based document summarization.
In this paper, we present \textit{QBSUM}, a high-quality large-scale dataset consisting of 49,000+ data samples for the task of Chinese query-based document summarization.
We also propose multiple unsupervised and supervised solutions to the task and demonstrate their high-speed inference and superior performance via both offline experiments and online A/B tests.
The QBSUM dataset is released in order to facilitate future advancement of this research field.
\end{abstract}


\end{frontmatter}


\section{Introduction}
\label{sec:intro}


Query-based document summarization aims to produce a compact and fluent summary of a given document which answers or is relevant to the search query that leads to the document.
Extracting or generating query-based document summarization is a critical task for search engines \cite{wang2007learning,sun2005web}, news systems \cite{liu2017growing}, and machine reading comprehension \cite{he2017dureader,choi2017coarse,wang2019evidence}.
Summarizing a web document to answer user queries helps users to quickly grasp the gist of the document and identify whether a retrieved webpage is relevant, thus improving the search efficiency.
In practical applications, query-based document summarization may also serve as an important upstream task of machine reading comprehension, which aims to generate an answer to a question based on a passage. The summary can be considered as evidence or supporting text from which the exact answer may further be found.

Existing research on text summarization can be categorized into generic text summarization and query-based text summarization.
Generic text summarization produces a concise summary of a document, conveying the general idea of the document \cite{carbonell1998use}, \cite{mcdonald2007study}, \cite{gillick2009scalable}, \cite{erkan2004lexrank}.
A number of diverse datasets have been developed, including Gigaword \cite{graff2003english}, New York Times Corpus \cite{sandhaus2008new}, CNN / Daily Mail \cite{hermann2015teaching}, and NEWSROOM \cite{grusky2018newsroom}, etc.
In contrast, query-based document summarization must produce a query-biased result answering or explaining the search query while still being relevant to the document content, which is a more challenging task.
Datasets created to date for this task are often of a tiny scale such as DUC 2005 \cite{dang2005overview}, or built upon human-crafted rules using web-crawled information \cite{nema2017diversity}.
To the best of our knowledge, there is not yet any publicly available dataset developed for Chinese query-based document summarization.


To address the demand for a large and high-quality query-based document summarization dataset and to facilitate the advancement of related research, in this paper, we present the \textbf{Q}uery-\textbf{B}ased document \textbf{SUM}marization (\textbf{QBSUM}) dataset, which consists of $\langle$\textit{query, document, summarization}$\rangle$ tuples, where the summarization to each query-document pair is a collection of text pieces extracted from the document (with an example shown in Fig.~\ref{fig:data}) labeled by five professional product managers and one software engineer in Tencent. QBSUM contains more than 49,000 data samples on over 49,000 news articles, where queries and documents are extracted based on queries and search logs of real-world users in QQ browser that serves over 200 million daily active users all around the world. 

Beside the large-scale and high-quality data collected from real-world search queries and logs, the QBSUM dataset is also created with considerations to various quality measurements, including relevance, informativeness, richness, and readability. 
Firstly, the selected summary must be relevant to the query as well as the major focus of the corresponding document. 
If a query can directly be translated to a question, the summary shall contain the answer to the question if applicable, or provide information that is helpful for answering the question.
Furthermore, the summary shall convey rich and non-redundant information related to the query.
Lastly, while being concise, the natural language summary must also be fluent for readability purposes---simple concatenation of several text pieces may not always serve the purpose.

To tackle the task of query-based document summarization, we design and implement three solutions: i) an unsupervised ranking model based on relevance defined in \cite{peyrard2019simple}; ii) an unsupervised ranking model based on a range of features; and iii) a query-based summarization model based on BERT \cite {devlin2018bert}.
We evaluated the performance and inference efficiency of different models on the QBSUM dataset, and compared with multiple existing query-based summarization baseline methods.
Our best model achieves a BLEU-4 score of 57.4\% and ROUGE-L score of 73.6\%, which significantly outperforms the baseline methods.

Based on the QBSUM dataset, we trained and deployed our query-based document summarization solution into QQ browser and Mobile QQ, two real-world applications involving more than $200$ million daily active users all around the globe.
Our solution currently serves as the core summarization system in these commercial applications for extracting and presenting concise and informative summaries based on user queries, to improve the search effectiveness and efficiency in these applications.
Furthermore, we conducted large-scale online A/B tests on more than 10 million real-world users in QQ browser mobile app.
The experimental results suggest that our model is able to improve the search results with web document summaries conforming to user queries and attention. 
The Click-Through Rate (CTR) increased by 2.25\% after our system was incorporated into the search engine.
To facilitate advancements in related research and tasks, we open source the QBSUM dataset \footnote{\url{https://www.dropbox.com/sh/t2cp7ml1kb8ako0/AADmS2RMfJvLbukyQbb08CGGa?dl=0}} and will later release our code for experiments as well. 

\section{Dataset Collection and Analysis}
\label{sec:data}

\begin{figure*}[t]
\includegraphics[width=\textwidth]{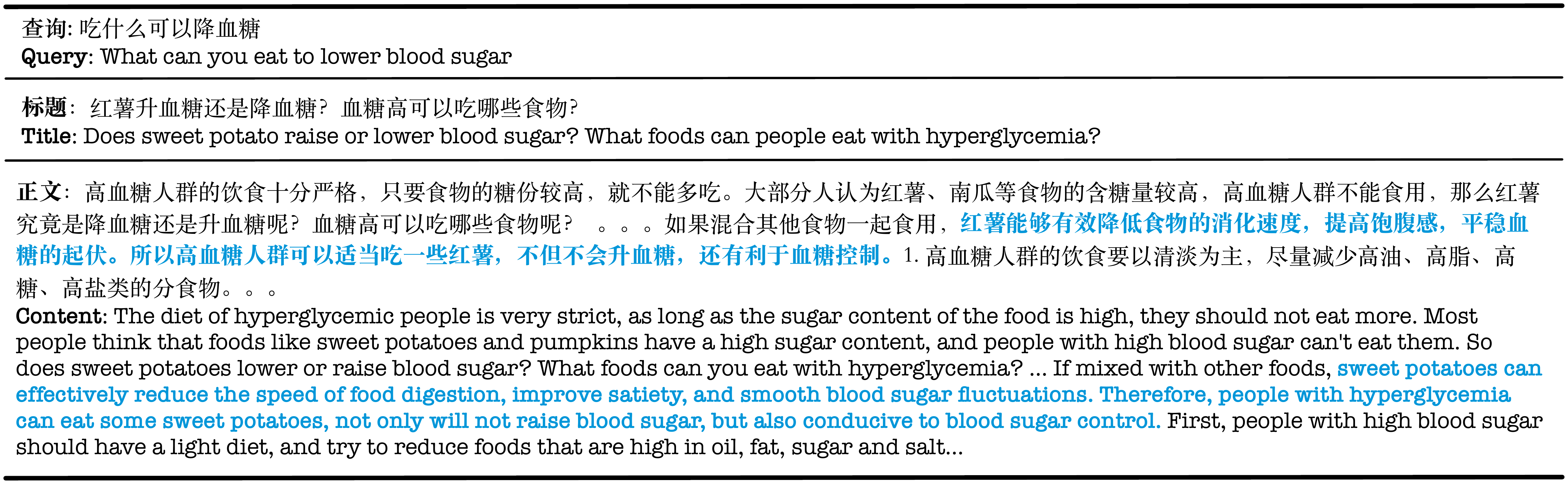}
\caption{An example of the query-doc-summary tuple in our QBSUM dataset. The summary is shown in blue text.}
\label{fig:data}
\end{figure*}

In this section, we introduce the procedures of constructing the QBSUM dataset, and analyze its specific characteristics. Fig.~\ref{fig:data} gives an data example presented in the dataset.

\subsection{Data Collection}
\label{sec:collect}

\textbf{Query and document curation}. We collect the queries and documents in our dataset from
 Tencent QQ Search (\url{http://post.mp.qq.com}), 
one of the most popular Chinese news websites.

We retrieve a large volume of queries and its top clicked articles posted between June 2019 and September 2019.
For each article, we perform text segmentation with punctuations \begin{CJK}{UTF8}{gbsn}[[，？！。]\end{CJK}, and filter out short articles with less than 15 text segments, as well as long articles with more than 10 paragraphs.
The retrieved samples covers a wide range of topics such as recent events, entertainment, economics, etc.
After that, we tokenize each query and article, and filter out samples in which the article and the query do not possess token overlaps.
Table \ref{tab:statistics} provides the key data statistics and a comparison with two existing query-based summarization dataset.
The QUSUM dataset has a vocabulary size of 103,005. The average length (number of words) of the queries, articles, and summaries are 3.82, 378.57 and 53.08, respectively. Comparing to DUC2005 and DUC2006, QBSUM holds a scale two orders magnitude larger and covers a far wider range of subjects, which fills the pressing need of a large-scale dataset in the research field of query-based summarization.

\textbf{Summary annotation}. Next, we annotate the summary of each collected query-document pair.
The document is denoted as $D$ consisting of $m$ text pieces $\mathcal{S} = \{S_1, S_2, \cdots, S_{m}\}$. And a query $Q$ is a sequence of $n$ query words $q_1q_2\cdots q_{n}$.
The annotators are asked to select at most $k=10$ text pieces $S_i \in \mathcal{S}$ that are most relevant to $Q$ and convey valuable information to users.
The summary $Y$ is constructed by concatenating the selected text pieces in the order they present in the document.
In terms of quality control of the annotated summary, we refer to the following four criteria:
\begin{itemize}
	\item \textbf{Relevance}. The summary must be semantically relevant to the user query. They may contain the keywords of the query.
	\item \textbf{Informativeness}. The summary shall include the answer to the query and provide valuable information or explanations if such information is available in the document. Supposedly, it should reduce user's uncertainty about the query.
	\item \textbf{Richness}. The summary shall contain diverse and non-redundant information that is relevant to the query. 
	\item \textbf{Readability}. The summary shall be consistent for a good readability to real-world users. 
\end{itemize} 

As an aid to annotators, for each text piece, we estimate its importance by TextRank \cite{mihalcea2004textrank} and indicate whether it contains at least one content word in the query.
The detailed procedure of annotation is introduced as follows:
\begin{enumerate}
  \item Check whether the document $D$ is relevant to the query $Q$. Discard this sample if not.
  \item Check the text pieces that share the same content words with $Q$, and read over the context sentences they belong to. Find out the most relevant and important text pieces with respect to $Q$.
  \item If the selected text pieces in the document are not consistent or fluent, decide whether to expand them to include their neighboring text pieces. A neighboring chunk is selected if it helps to improve the readability of summary $Y$ and contains relevant and non-redundant information to $Q$.
  Each selected text piece $S_i$ is expanded with no more than $3$ neighboring chunks.
  The overall number of text pieces in $Y$ shall not exceed $k$, and the maximum number of words is set to $70$.
\end{enumerate}

\textbf{Annotation Cost}.
Each worker needs to extract text segments related to the query from documents composed of 49 segments in average, according to their relevance and informativeness. At the same time, the adjacent text pieces need to be carefully selected, such that the requirements of richness and readability can be fulfilled.
Due to the high complexity of the annotation process, the time cost of data annotation is relatively expensive with the average cost for annotating a single piece of data being 115 seconds.

\textbf{Quality control}.
Our workers comprise 5 professional product managers and 1 software engineer in Tencent.
Each worker needs to take an annotation examination to ensure their intact understanding about how to annotate and the ability of extracting qualified summaries according to the multi-aspect criteria.
Three rounds of summary annotation are conducted.
First, five workers will annotate different groups of the dataset individually.
Second, each worker will review $5\%$ of data sampled from the groups annotated by other workers.
Third, an expert reviewer will review all the data examples created by workers.
This loop is repeated to make sure the accuracy (judged by the expert reviewer) is above 95\%.

\begin{table}[tb]
\centering
\small
  \begin{tabular}{l|ccc}
    \toprule
    Datasets & QBSUM & DUC2005 & DUC2006 \\
    \midrule
    \# source sentences & $2,175,639$ & $14,410$ & $18,794$  \\
    \# summary sentences & $105,751$ & $4,242$ & $5,037$  \\
    avg \#characters/query  & $7.1$ & $-$ & $-$  \\
    avg \#characters/summary  & $41$ & $250$ & $250$  \\
    \#unique queries & 16,250 & $-$ & $-$ \\
    \#unique documents & 43,762 & $-$ & $-$ \\
    \#doc-summary-query & 49,535 & $-$ & $-$ \\
    avg \#words/document & 378.57 & $-$ & $-$ \\
    avg \#words/query  & 3.84  & $-$ & $-$ \\
    avg \#words/summary & 53.08 & $-$ & $-$ \\
    \bottomrule
  \end{tabular}
  \caption{The statistical information of QBSUM and compare it with the DUC datasets.}
  \label{tab:statistics}
\end{table}

\subsection{Data Analysis}
\label{sec:analysis}

\textbf{Statistical analysis}.
Table \ref{tab:statistics} describes the statistical information of QBSUM and compares it with the DUC2005 and DUC2006 datasets. As the DUC datasets are not publicly available, we can only collect the shown information from paper \cite{dang2005overview}.
We can see that QBSUM is two orders of magnitude larger than DUC2005 and DUC2006.
The average number of characters in a summary in QBSUM dataset is $41$ which is notably lower than the average 250 characters in DUC2005 and DUC2006. The reason of this difference is traced back to the characteristics of Chinese language where a Chinese word is usually made up of 1 to 4 characters which is far less than English words. 

\begin{figure}[t]

                        \centering
                \includegraphics[width=0.6\textwidth]{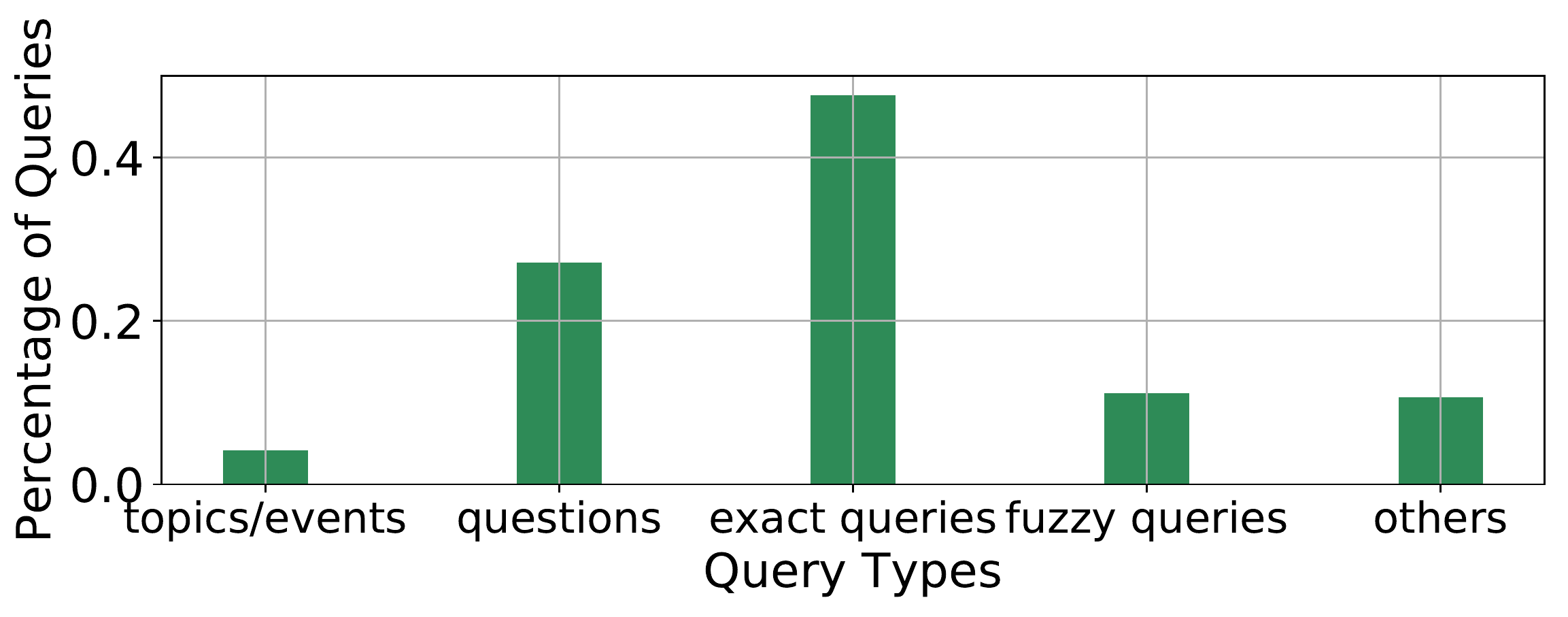}
                \caption{The distribution of query types in QBSUM.}
                \label{fig:queryType}

\end{figure}

\begin{figure}
\centering
\includegraphics[width=0.6\textwidth]{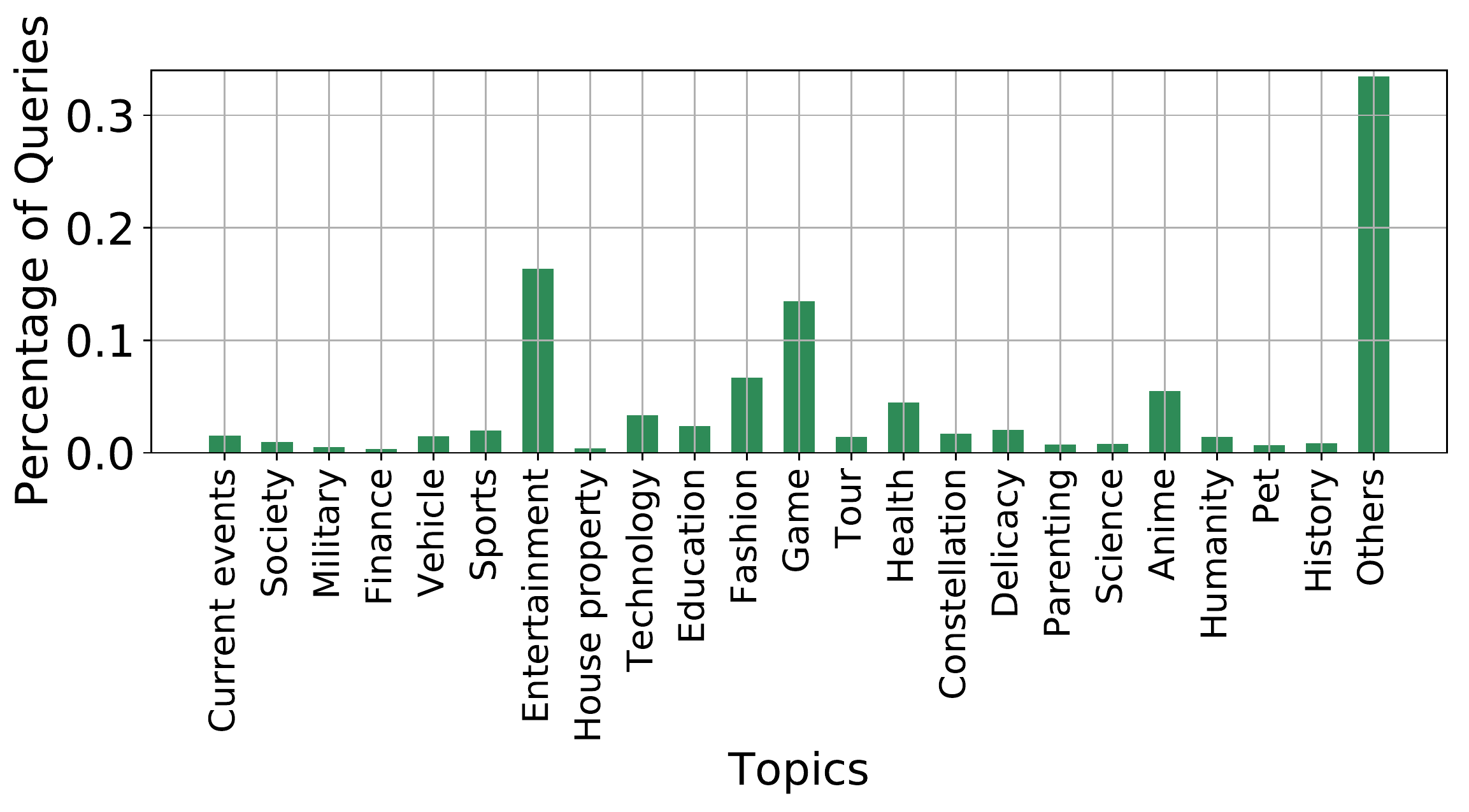}
\caption{Topic distribution in QBSUM.}
\label{fig:queryTopic}

\end{figure}

\textbf{Query type distribution}. The DUC 2005 and 2006 tasks are question-focused summarization tasks which concentrate on summarizing and integrating information extracted from multiple documents to answer a question.
In contrast, our QBSUM dataset contains a diversity of queries produced by real-world users, which can be generally categorized into 5 types:
\begin{itemize}
	\item \textbf{Hot topics/events}. This type of queries is focused on recent hot topics or events. The corresponding summaries are mainly descriptions about the event development, location, related entities, time and so forth.
	\item \textbf{Questions}. This category covers user questions such as ``what universities are recommended for a GMAT score of 570''. The summaries are the answers or evidence sentences to the questions.
	\item \textbf{Exact queries}. Such queries are about some specific concepts or entities, such as ``country houses under 100 square meters in Shenzhen''. The summaries contain useful information that are relevant to the queries.
	\item \textbf{Fuzzy queries}. This type of queries includes fuzzy or subjective concepts or questions, such as ``top ten fuel-efficient cars''. The corresponding summaries provide information of relevant entities or the explicit answers to the queries.
	\item \textbf{Unclear/incomplete queries}. For this type, users may be not clear about how to express their intention. Therefore, the summaries cannot give specific answers to the queries.
\end{itemize}
We sample 500 queries from the QBSUM dataset and manually check their query types where the data distribution is shown in Fig.~\ref{fig:queryType}. The largest partition of the query type in QBSUM belong to exact queries and questions while the percentage of topics/events are relatively small, due to the limited occurrences of hot topics/events in our daily lives. However, the Click-Through Rate (CTR) of hot topics/events are actually quite high.

\textbf{Topic distribution}.
The QBSUM dataset covers queries and documents in a wide range of topics. Fig.~\ref{fig:queryTopic} also presents the distribution of document topics in which we can see that a large portion of documents are discussing topics about entertainments, games, fashion or anime.

\section{Methods}
\label{sec:model}

We developed three models for query-based document summarization including two fast unsupervised methods and a high-performance supervised model based on pre-trained BERT \cite{devlin2018bert}.


\subsection{Relevance-based Summarization} 
\label{sub:relsum}
Intuitively, in query-based summarization, a summary is profitable for a user, if it yields effective knowledge of the field which the query is focused on. Formally, \textit{relevance} measures the information loss between the distribution of a summary $Y$ and the requested knowledge field (i.e., the query $Q$) \cite{peyrard2019simple} which is defined via the cross-entropy $CE(Y,Q)$:
\begin{align}
Rel(Y, Q) = \sum_{w_i \in Y\cup Q}\mathbb{P}_Y(w_i)\cdot\log (\mathbb{P}_Q(w_i)), \label{eq:rel}
\end{align}
where $w_i$ is a character appearing in $Y$ or $Q$.
We propose to maximize $Rel(Y, Q)$ or $Rel(Q, Y)$ under the limitation of summary length. Specifically, we test the following methods based on relevance:

\begin{itemize}
	\item \textbf{Rel-YQ-top6}. Iteratively select the top $N$ text pieces of document $D$ according to $Rel(Y, Q)$. The iteration stops until $N = 6$ or the number of characters in $Y$ reaches or exceeds $70$ for the first time.  
	\item \textbf{Rel-YQ-top3-expand}. Iteratively select the top $3$ text pieces according to $Rel(Y, Q)$. Expand each text piece in order to include its preceding and following text piece until the number of characters in $Y$ reaches or exceeds $70$. 
	\item \textbf{Rel-QY-top3-expand}. Similar with the above method, except that we replace $Rel(Y, Q)$ with $Rel(Q, Y)$.
	\item \textbf{Rel-QY-top2-expand}. Similar with the above method, except that we replace $N = 3$ with $N = 2$.
\end{itemize}

\begin{figure*}[t]
\includegraphics[width=\textwidth]{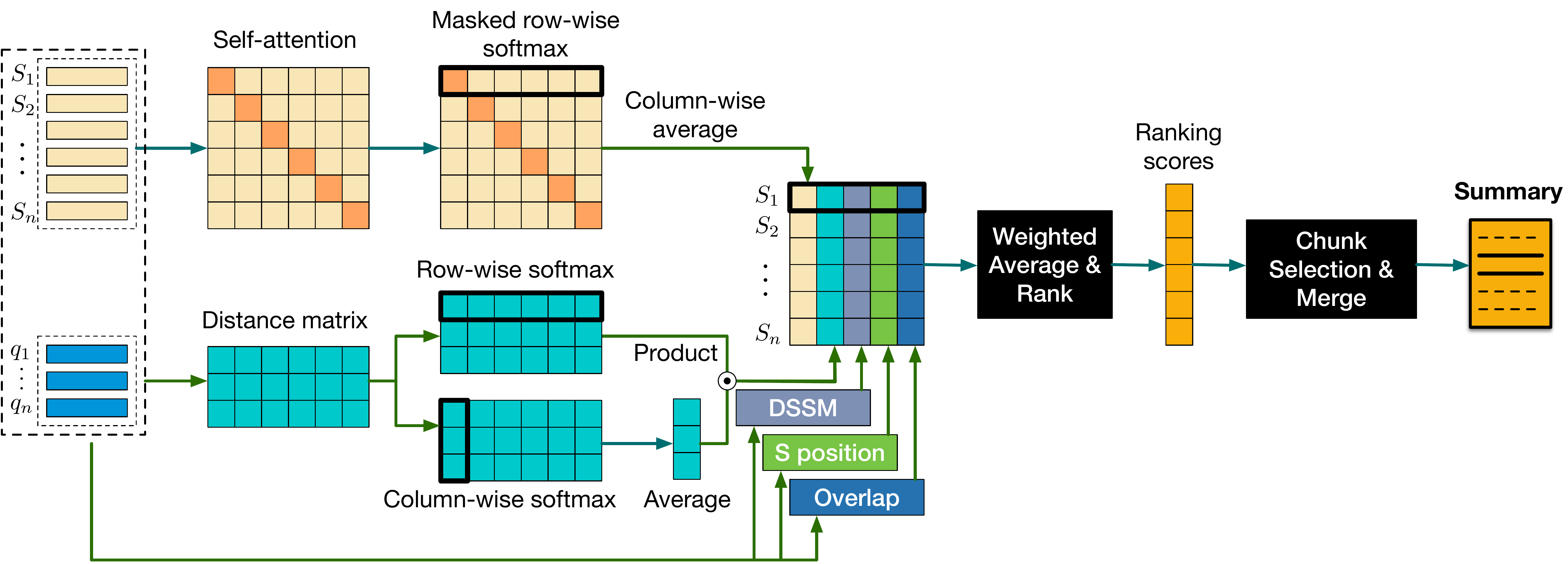}
\caption{The architecture of our online query-based summarization model.}
\label{fig:model1}
\end{figure*}

\subsection{Ranking with Dual Attention}
\label{sub:rankdualatt}
Fig.~\ref{fig:model1} presents the architecture of our online unsupervised summarization model.
We first acquire the representations of the query and the document with pre-trained word embeddings. We represent the query by a collection of embedding vectors of the query words it contained. And the document representation is formed by its text pieces, where the embedding vector of each text piece is computed by taking an average of its word vectors.
Our model then exploits the derived representations to compute various feature scores with respect to different dimensions of the samples. The main features shown in Fig.~\ref{fig:model1} include the following:
\begin{itemize}
	\item \textbf{S-D self-attention}. This feature measures the importance of each text piece $S \in \mathcal{S}$ to $D$ by interacting it with the other remaining text pieces in the document and calculating the self-attention scores of $S \in \mathcal{S}$. We follow \cite{vaswani2017attention} for the computation of attention scores. A summary should disclose important information of the document.
	\item \textbf{S-Q co-attention}. This feature measures the relevance of each sentence $S \in \mathcal{S}$ to $Q$ by calculating the co-attention scores between $Q$ and itself. We suppose the summary should be highly relevant to the query.
	\item \textbf{S-Q semantic matching (DSSM)}. This feature measures the semantic relevance between $S$ and $Q$ by a Deep Structured Semantic Matching (DSSM) model \cite{huang2013learning}. The DSSM model is trained in a pair-wise setting with $0.2$ billion query-title pairs from Sougou search engine, where we utilize the queries and the titles of top 1 clicked documents as positive examples, and sample from random combinations of query-title pairs to construct negative samples.
	\item \textbf{S position}. It indicates the sequential order of $S$ in $D$. We apply min-max normalization on the sequential number to calculate the position score.
	\item \textbf{S-Q overlap}. It measures the overlap between $S$ and $Q$ on word level and is estimated by $\frac{n_o}{n}$, where $n_o$ is the number of overlapping query characters in $S$, and $n$ is the total number of query characters.
\end{itemize}
In our implementation, we use different weights to combine the above feature scores where the weights are tuned as hyper-parameters, and derive an unsupervised model that ranks the scores of each $S$ belonging to the output summary $Y$. The feature weights can also be learned by simple logistic regression resulting in a supervised model. 

Next, we expand and combine the ranked text pieces to extract a query-based summarization $Y$. Algorithm~\ref{alg:model1} presents the detailed steps to extract the text pieces belonging to $Y$.
We iterate through each candidate text piece $S_{p_i}$ and expand it by its preceding and following text pieces (if available) to get $C_{p_i}$.
We estimate the redundancy of $C_{p_i}$ by the ratio of overlapping bi-grams between $C_{p_i}$ and $S_{p_i}$.
We discard $C_{p_i}$ if the redundancy reaches or exceeds certain threshold (we use 0.5). Otherwise, $C_{p_i}$ will be appended as a part of $Y$. We repeat this step until the length (number of words) of $Y$ is larger than a threshold (we use $70$).

\begin{algorithm}[t]
\KwIn{a sequence of sorted text pieces $\mathcal{S} = \{S_{p_1}, S_{p_2}, \cdots, S_{p_m}\}$, where $p_i$ is the position of $S_{p_i}$ in a document $D$.}
\KwOut{Summary $Y = \{S_{y_1}, S_{y_2}, \cdots, S_{y_{|Y|}}\}$.}

\begin{algorithmic}[1]
\STATE $Y \leftarrow \varnothing$\;
\FOR{each $S_{p_i} \in \mathcal{S}$}
    \STATE $C_{p_i} = [S_{p_i - 1}]S_{p_i}[S_{p_i + 1}]$\;
    \STATE $Redundancy = \frac{Count(Bigram(C_{p_i} \cap S_{p_i}))}{Count(Bigram(C_{p_i} \cup S_{p_i}))}$\;
    \IF {$Redundancy \geq 0.5$}
        \STATE Continue\; 
    \ENDIF
    \STATE $Y = Y \cup C_{p_i}$\;
    \IF {The length of $Y$ reaches threshold}
        \STATE Break\;
    \ENDIF
\ENDFOR
\STATE Sort the text pieces in $Y$ by their positions.
\end{algorithmic}
\caption{Greedy Sentence Selection}
\label{alg:model1}
\end{algorithm}

\begin{figure}[t]
\centering
\includegraphics[width=0.75\textwidth]{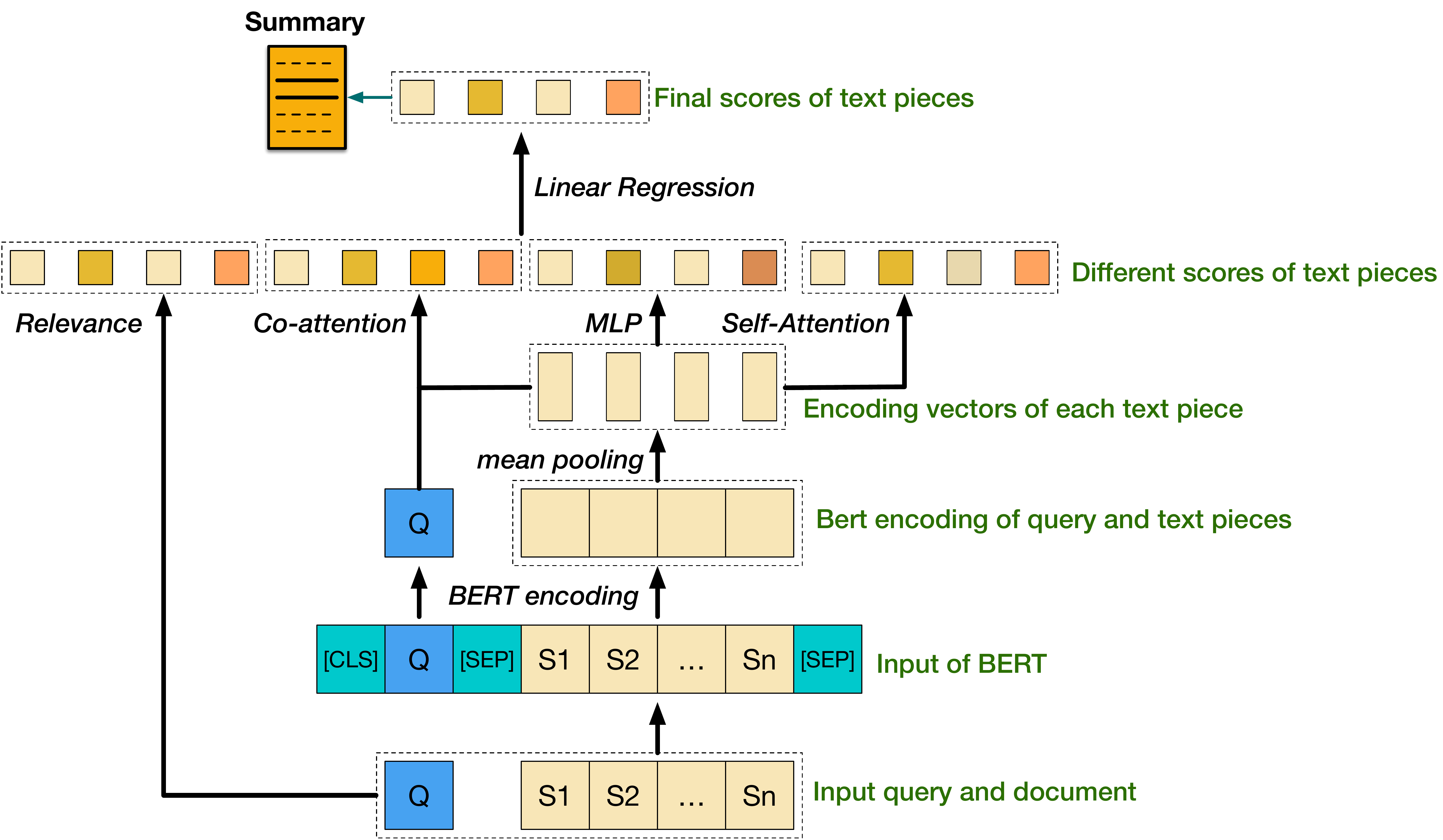}
\caption{The architecture of our BERT-based query-based summarization model.}
\label{fig:modelBert}
\end{figure}

\subsection{Query-based Summarization based on BERT}
Large-scale pre-training models such as BERT \cite{devlin2018bert} have dramatically advanced the performance on a wide range of NLP tasks.
We propose a simple BERT-based model with a pre-trained BERT of 110M parameters as encoder, and perform $S$-$Q$ text-pair classification to determine whether each text piece $S$ belongs to the summary $Y$. The architecture of the model is shown in Fig.\ref{fig:modelBert}.

In order to distinguish the query and the document in the input of BERT, we use a $[CLS]$ token at the start of the input followed by the query tokens, then a $[SEP]$ token followed by tokens of document, and another $[SEP]$ token is appended at the end of the document. 
Each token of the input is represented by the summation of its token embedding, segment embedding, and position embedding and is fed to the BERT model to obtain its encoded BERT representation. 
The self-attention mechanism adopted in BERT allows the learning of correlation between any two tokens in the document or the query. Hereby, the query information is also transmitted to the document representations.

Then, a mean-pooling layer is adopted to collect the sentence vector representation of each text piece in the document by taking the average among all the token vectors within the text piece. 
Furthermore, on top of the derived BERT sentence representations, a transformer layer is added to model the correlation among text pieces in the document and produce document-aware sentence representations, which are then fed through a linear projection layer that transforms the representation into a scalar score.


Inspired by Rel-QY in section \ref{sub:relsum} and Rank-DualAttn in section \ref{sub:rankdualatt}, we add several modules in addition to the BERT prediction including: i) a relevance module which computes the relevance between a given text piece and the query using eq.\ref{eq:rel}; ii) an S-D self-attention module which measures the importance of each text piece regarding to the document, note that we use the derived BERT representation instead of pre-trained word vectors in Rank-DualAttn; iii) an S-Q co-attention module which calculates the co-attention scores between the query and a text piece with BERT representations. The computed scores are then concatenated together with the BERT prediction score to make the final prediction to decide whether a text piece belongs to the output summary.




\section{Experiments}
\label{sec:eval}

In this section, we evaluate our approaches for query-based document summarization and make comparisons with multiple baselines. We also demonstrate the benefits QBSUM brought to search engines through large-scale online A/B testing.

\subsection{Experimental Setup}
\label{subsec:setup}

\subsubsection{Baseline methods} 
We compare our model with three baseline methods, Textrank-DNN, tfidf-DNN, and MDL \cite{litvak2017query}.

In \textbf{Textrank-DNN} and \textbf{tfidf-DNN}, an undirected graph is created from the document with nodes being the semantic embeddings of text pieces and edges being cosine similarities with the query. The difference is that Textrank-DNN uses Textrank algorithm \cite{mihalcea2004textrank} to compute the weights of nodes where tf-idf algorithm is used in tfidf-DNN.

\textbf{MDL} first selects frequent word sets related to the give query, then extracts summaries by selecting sentences that best cover the sets. 

For our Bert-QUSUM model, we perform ablation analysis to study the effectiveness of different modules of our proposed model and evaluate the following versions:

\begin{itemize}
\item \textbf{Bert-QBSUM (no relevance)}. In this variant, we keep the dual attention modules and the transformer sentence encoder, and remove the relevance module.
\item \textbf{Bert-QBSUM (no self-attention)}. This model variant does not contain the self-attention module.
\item \textbf{Bert-QBSUM (no co-attention)}. The co-attention module is not contained in this variant.
\item \textbf{Bert-QBSUM (no transformer encoder)}. The transformer sentence encoder is removed so that the output of mean-pooling layer is used as representation of text pieces to produce the BERT prediction scores. 
\item \textbf{Bert-QBSUM}. This is the complete version of our BERT-based summarization model.
\end{itemize}

\begin{table*}[!tb]
\centering
\small
  \resizebox{\textwidth}{!}{
  \begin{tabular}{c|cccc|ccc|c}
    \toprule
    \multirow{2}{*}{Models} & \multicolumn{4}{c}{BLEU} & \multicolumn{3}{c}{ROUGE} & Inference Time (s)\\ 
    & 1 & 2 & 3 & 4 & 1-recall & 2-recall & L-recall & (2,120 samples)\\
    \midrule
     TFIDF-DNN & $41.43$ & $36.38$ & $33.34$ & $31.24$ & $43.92$ & $33.09$ & $37.04$ & $35$ \\
    MDL & $36.48$ & $29.02$ & $24.99$ & $22.41$ & $37.64$ & $23.10$ & $29.87$ & $79$ \\
   TextRank-DNN & $42.89$ & $36.84$ & $33.33$ & $30.96$ & $44.23$ & $32.70$ & $36.67$ & $1,134$ \\
   \midrule
   Rel-YQ-top6 & $39.76$ & $34.16$ & $30.77$ & $28.30$ & $41.48$ & $29.41$ & $33.58$ & $30$\\
   Rel-YQ-top3-expand & $47.15$ & $40.79$ & $36.81$ & $33.85$ & $56.08$ & $41.43$ & $43.67$ & $20$ \\
   Rel-QY-top3-expand & $44.93$ & $39.09$ & $35.49$ & $32.85$ & $61.84$ & $46.01$ & $47.54$ & $20$ \\
   Rel-QY-top2-expand & $47.73$ & $41.69$ & $37.93$ & $35.16$ & $53.64$ & $39.41$ & $42.67$ & $15$ \\
   Rank-DualAttn & $50.78$ & $45.68$ & $42.33$ & $39.77$ & $58.68$ & $45.96$ & $56.17$ & $762$ \\
   \midrule
   Bert-QBSUM (no relevance) & $60.99$ & $58.40$ & $56.88$ & $55.83$ & $72.89$ & $67.03$ & $71.95$ & $12,944$ \\
   Bert-QBSUM (no co-attention) & $61.64$ & $59.03$ & $57.52$ & $56.48$ & $73.17$ & $67.34$ & $72.23$ & $11,391$ \\
   Bert-QBSUM (no self-attention) & $62.20$ & $59.62$ & $58.13$ & $57.12$ & $72.42$ & $66.55$ & $71.46$ & $11,236$ \\
   Bert-QBSUM (no transformer encoder) & $62.03$ & $59.52$ & $58.05$ & $57.03$ & $73.67$ & $67.92$ & $72.62$ & $10,767$ \\
   Bert-QBSUM & $\mathbf{62.40}$ & $\mathbf{59.91}$ & $
   \mathbf{58.45}$ & $\mathbf{57.45}$ & $\mathbf{74.69}$ & $\mathbf{68.91}$ & $\mathbf{73.58}$ & $14,000$ \\
    \bottomrule
  \end{tabular}
  }
  \caption{Compare the performance of different approaches based on the QBSUM dataset.}
  \label{tab:experiments}
\end{table*}

\subsubsection{Metrics}
We use ROUGE and BLEU to evaluate model performance. 

\textbf{ROUGE} \cite{lin2004rouge} measures the quality of a summary by comparing it to reference summaries by counting the number of overlapping units. In our experiment, we use n-gram recall ROUGE-N with $N=1,2$ and ROUGE-L based on Longest Common Subsequence (LCS) statistics.

\textbf{BLEU} \cite{papineni2002bleu} measures precision by how much an n-gram text in prediction sentences appear in reference sentences at the corpus level. BLEU-1, BLEU-2, BLEU-3, and BLEU-4, use 1-gram to 4-gram for calculation, respectively.

\subsubsection{Implementation details}
For our experiments, we utilize a part of the QBSUM dataset with around 10,000 data samples as the rest of the data was collected after the experiments were conducted. 
The dataset is split into a train set, a evaluation set and a test set consisting of $8787$, $1099$ and $1098$ samples respectively. 
All performances are reported on the test set.
In the implementations of our unsupervised models and the baseline models, we use a 200-dimensional Chinese word embedding trained using w2v \cite{mikolov2013efficient} on 2 billion queries. The maximum length of generated summaries is set to 70 Chinese words. 

\subsection{Results and Analysis}
\label{sec:result}

Table~\ref{tab:experiments} summarizes the performance of all the compared methods on the QBSUM dataset.
As the inference speed of models are critical to real-world applications such as search engines or recommender systems, we also report inference time of different methods on a set of 2,120 samples in QBSUM.
Among all methods, our unsupervised methods achieve both better performance and faster inference when compared with baseline methods. And our supervised BERT-based models produce the best results.

By observing the results, we can see that the Bert-QBSUM model obtains the best performance in terms of all performance metrics as shown in table \ref{tab:experiments} and achieves a huge performance gain compared with previous models. 
The success of our Bert-QBSUM model can be attributed to the powerful encoding ability of BERT as well as the combination of different feature modules in our model. 
By carefully examining the performance of different variants of QB-SUM, we can see that when the co-attention and relevance are removed, the performance drops notably, from which we can conclude that modeling the relevance between each text piece and the query plays an essential role in query-based summarization. 
The transformer encoder and the self-attention module are also beneficial to the model performance, as they are capable of modeling the correlation between text pieces in documents on sentence level where BERT only models the information on token level.

However, Bert-QBSUM is quite slow in inference than other methods due to its large number of model parameters.
The relevance-based summarization model variants achieve the best inference speed because they are unsupervised methods and require less calculations during inference. 
As a trade-off between inference speed and performance, our online Rank-DualAttn model achieves better performance within acceptable inference time, making it the most suitable model for real-world applications.

\begin{figure}[tb]
\centering
\includegraphics[width=4.0in]{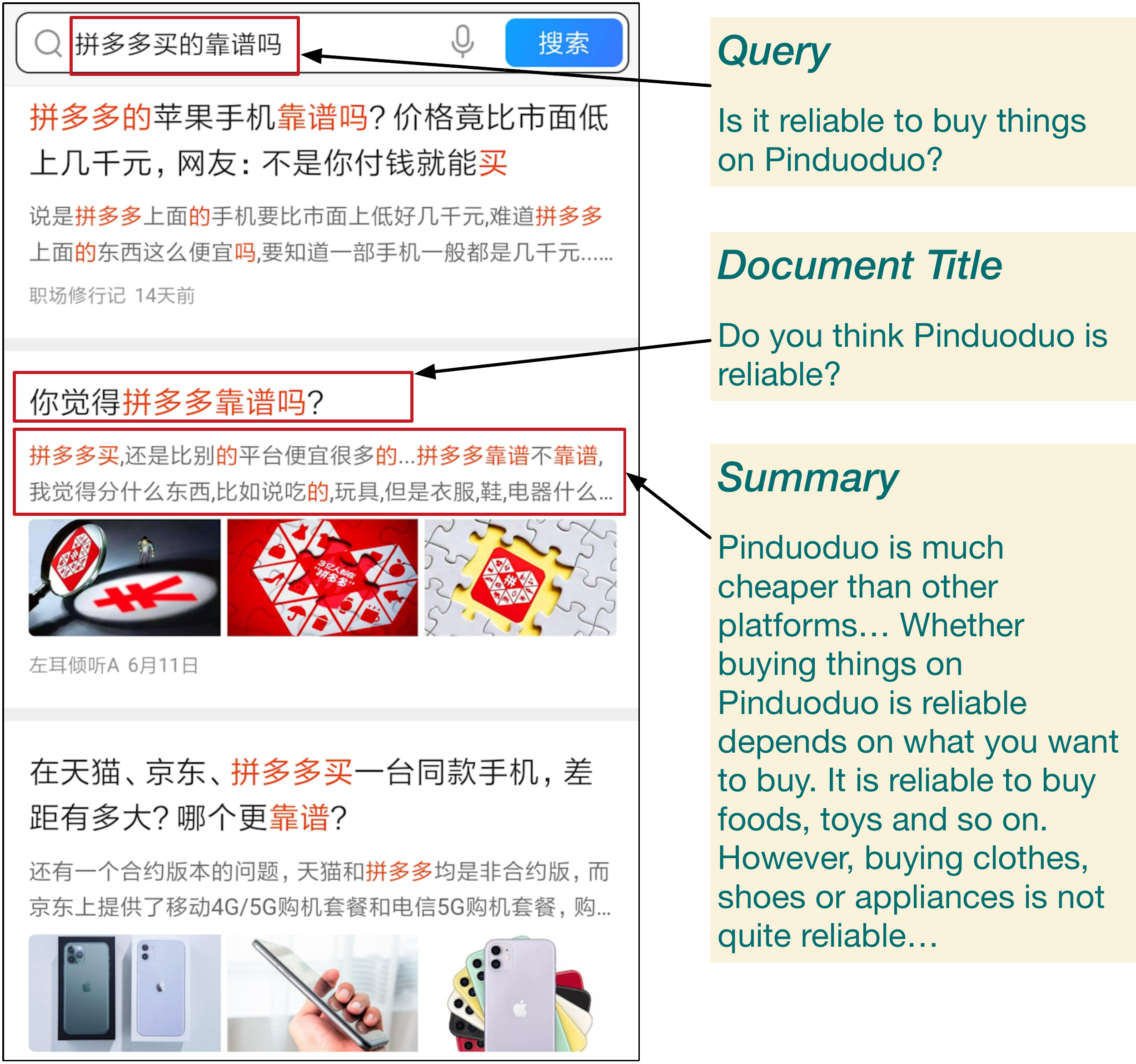}
\caption{Example of query-based document summarization in QQ browser.}
\label{fig:DocTagExample}
\end{figure}

The Rank-DualAttn model is deployed into QQ browser that involves more than 0.2 billion daily active users all around the globe.
Fig.~\ref{fig:DocTagExample} demonstrates its effectiveness through an example. When a user inputs a query ``Is it reliable to buy things on Pinduoduo?'', our model is able to extract a summary from the top clicked documents which contains the answer to the query and relevant explanations and details. 
And the performance gain achieved proved the effectiveness of our designed features.

We performed error analysis on 100 summaries sampled from our applications.
Based on our observation, the errors can be mainly divided into two groups: i) \textit{low relevance}, which means the summary are not quite relevant to the query, and ii) \textit{incomplete information}, i.e., the summary only contains an incomplete part of the key information relevant to the query.

The main cause of the errors is due to the fact that the semantic relevance estimation between query and document sentences is inaccurate. As supplementary, the model emphasizes more on low-level features such as keyword overlapping or textual similarities.
For example, given a query ``What is the order of the Chinese zodiac signs'', the model is not capable of capturing the hidden information that ``Chinese zodiac signs'' represent 12 kinds of animals. As a result, the extracted summary usually contains the keyword   ``Chinese zodiac signs'', but not exactly the answer of ``the order of the Chinese zodiac signs''.
Our analysis shows that low relevance errors often happen for the type of exact queries, and incomplete information errors emerge most frequently for question type queries.

\subsection{Online A/B Testing for Query-based Document Summarization}

We perform large-scale online A/B testing to show how query-based document summarization helps with improving the performance of searching in real world applications.

For online A/B testing, we split users into buckets where each bucket contains 5 million users. We first observe and record the activities of each bucket for 3 days based on the following metrics:
\begin{itemize}
  \item \textbf{Click-Through Rate (Global-CTR)}: the ratio of users who clicked on any of the search results to the total users who received the results.
  \item \textbf{Top 1 Click-Through Rate (Top1-CTR)}: the ratio of users who clicked on the top 1 search result to the total users who received the result.
  \item \textbf{Top 2 Click-Through Rate (Top2-CTR)}: similarly, here we use top 2 results.
  \item \textbf{Top 3 Click-Through Rate (Top3-CTR)}: similarly, here we use top 3 results.
  \item \textbf{Selection Rate}: the ratio of  user clicked queries to the total queries.
\end{itemize}
We then select two buckets with highly similar activities where our Rank-DualAttn query-based summarization model is utilized in one of the buckets while in the other buckets the first few sentences of the document are presented to users as the summarization.
We run our A/B testing for 3 days and compare the results on the above metrics.

\begin{table}[tb]
\centering
  \caption{Online A/B testing results.}
  \label{tab:ab-testing}
  \begin{tabular}{c|c}
    \toprule
    Metrics & Percentage Lift\\
    \midrule
    Selection Rate & +0.33\% \\
    Top1-CTR & +0.58\% \\
    Top2-CTR & +1.34\% \\
    Top3-CTR & +1.38\% \\
    Global-CTR & +2.86\% \\
    \bottomrule
  \end{tabular}
\end{table}

Table \ref{tab:ab-testing} shows the results of our online A/B testing. In the online experiment, we observe a statistically significant Global-CTR gain (2.86\%) when employing our Rank-DualAttn model. We also detect improvements on other metrics such as selection rate and top-N CTRs.
These observations prove that our online model for query-based summarization greatly benefits the search engine and grants users better experience by precisely capture their interested documents. 
With the help of query-based summarization, we can better capture relevant and helpful information in a document attractive to users and display the summaries before users click into the search results. Such summaries grant users the ability of quickly retrieve the core meaning of a document and locate the document of their true interest, hence improving both the effectiveness and the efficiency of the search engine.

\section{Related Work}
\label{sec:related}

\subsection{Existing Datasets}
In this section, we introduce existing datasets for document summarization and their characteristics.

\textbf{Generic Document Summarization} aims at compressing long documents into human readable short summaries that contain the most important information without specific focuses. 
In the past years, several large-scale summarization datasets have been introduced to accommodate the advance of this field.

Gigaword \cite{graff2003english} is a large-scale dataset containing more than 8 million documents from different newswire sources and corresponding headlines which are used as simulated summaries in prior work \cite{rush2015neural} \cite{chopra2016abstractive}. 
This compromise results in shorter summaries than most natural summary text. 
The New York Times Annotated Corpus \cite{sandhaus2008new} is a collection of over 1.8 million articles from the New York Times magazine between 1987 and 2007, with manually written and tagged summaries.
It has been used for both extractive document summarization \cite{li2016role} \cite{xu2019neural} and abstractive document summarization \cite{gehrmann2018bottom} \cite{celikyilmaz2018deep}.

CNN / Daily Mail question answering dataset \cite{hermann2015teaching} is originally introduced as a Cloze-style QA dataset, but also widely adopted on generic document summarization. It consists of CNN and Daily Mail articles, each associated with several bulletin point descriptions which are concatenated to form a summary.
\cite{grusky2018newsroom} presented the NEWSROOM dataset, consisting of 1.3 million articles and summaries written by their original authors and editors. LCSTS \cite{hu2015lcsts} is a Chinese short text summarization dataset collected from Weibo. It consists of over 2 million short texts with short summaries given by the author of each text. 

\textbf{Query-based Document Summarization}, compared with generic document summarization, highlights the points in the document relevant to the context of a query. It is of great value to question answering and search engines. However, due to the lack of datasets, this problem has drawn much less attention. 

DUC \footnote{https://www-nlpir.nist.gov/projects/duc/index.html} and CAT \footnote{https://tac.nist.gov/} have proposed several query-based summarization task in the past years
, where each summary is focused on a number of complex questions
However, they only provide a small test dataset which is far from satisfactory. 

Some researchers studying query-based summarization create datasets themselves by crawling from the web \cite{nema2017diversity} or generating queries from hand-crafted rules. However, such datasets often suffer from poor quality control and lack of data diversity. 

QBSUM contains more than 49,000 $\langle$\textit{query, document, summarization}$\rangle$, with considerations of various quality measurements, including relevance, informativeness, richness, and readability. 
To the best of our knowledge, QBSUM is the first large-scale high-quality Chinese query-based summarization dataset.

\subsection{Query-Based Summarization Methods}
Document summarization methods can be classified into extractive summarization which summarizes the document by extracting key words and phrases without modification, and abstractive summarization which generates new sentences to form a summary. 

Work on generic extractive summarization spans a large range of approaches. Maximum Marginal Relevance (MMR) \cite{carbonell1998use} is a widely known greedy approach. \cite{mcdonald2007study} \cite{gillick2009scalable} formulate this question as an Integer Linear Programming problem. Graph-based models also play a leading role in this field due to its ability to construct sentence relationships \cite{erkan2004lexrank} \cite{parveen2015topical} \cite{parveen2015integrating}. Recently, reinforcement learning methods have been applied \cite{narayan2018ranking}. 
For example, \cite{narayan2018ranking} conceptualize extractive summarization as a sentence ranking task and optimize ROUGE through an RL objective.

Many query-based summarizers are heuristic extensions of generic summarization methods by incorporating the information of the given queries. A variety of query dependent features were defined to measure the relevance, including TF-IDF cosine similarity \cite{wan2009graph}, WordNet similarity \cite{ouyang2011applying}, and word co-occurrence \cite{prasad2007iiit}, etc. \cite{cao2016attsum} proposed a joined attention model AttSum to meet the query need and calculate sentence weight.

\section{Conclusion}
\label{sec:conclude}





In this work, we introduce a new Chinese query-based summarization dataset called QBSUM, which to the best of our knowledge, is the first large-scale high-quality dataset in query-based summarization. QBSUM contains more than $49,000$ data samples collected from real-world applications, and is two magnitudes larger in scale than existing datasets such as DUC2005 and DUC2006. The QBSUM dataset is released and we hope that this dataset will promote future development of this research field. 

Additionally, we propose several supervised and unsupervised models which incorporate different properties of queries and documents including relevance, informativeness, and importance. The experiments on our QBSUM dataset demonstrate that our methods surpass other baselines.

In the future, we plan to further study the interactions among the queries, documents and summaries and develop abstractive methods on the QBSUM dataset. 




\bibliography{main}







\end{document}